\documentclass[10pt,twocolumn,letterpaper]{article}

\usepackage[pagenumbers]{cvpr}      
\definecolor{cvprblue}{rgb}{0.21,0.49,0.74}
\usepackage[pagebackref,breaklinks,colorlinks,allcolors=cvprblue]{hyperref}
\usepackage[table]{xcolor}
\usepackage{listings}
\usepackage{float}
\usepackage[most]{tcolorbox}
\tcbuselibrary{skins,breakable}
\usepackage{tikz}
\usepackage{pgfplots}
\usepackage{multirow}

\def\paperID{20194} 
\def\confName{CVPR}
\def\confYear{2026}

\title{Adapting Like Humans: A Metacognitive Agent with Test-time Reasoning}

\author{
Yang Li$^{1,}$\thanks{These authors contributed equally to this work.
$^\dagger$ Corresponding author.
$^{1}$Huawei Noah’s Ark Lab.
$^{2}$University of Oxford.
$^{3}$Tsinghua University.
$^{4}$University of Liverpool.
$^{5}$University College London.
} \and Zhiyuan He$^{1,*}$ \and Yuxuan Huang$^{4,*}$\and Zhuhanling Xiao$^{2}$  \and Chao Yu$^{3}$ \and Meng Fang$^{4}$ \and Kun Shao$^{1}$ \and Jun Wang$^{5,\dagger}$
}

\begin{document}
\maketitle
\begin{abstract}
Recent Vision-Language Models (VLMs) exhibit strong perceptual reasoning abilities, yet they often struggle to adapt efficiently when encountering novel tasks at test time. 
In contrast, humans leverage the metacognitive model with memory, enabling continuous strategy refinement through metacognitive control when faced with new challenges.
To bridge this gap, we propose metacognitive test-time reasoning (MCTR), a framework that equips models with the ability to learn, adapt, and improve during test time through metacognitive self-updating. 
Inspired by the dual structure of human metacognition, MCTR comprises meta-level and object-level VLM reasoning modules, each equipped with dedicated memory systems for hierarchical adaptive reasoning.
Specifically, MCTR consists of
(1) a meta-reasoning module which incrementally builds a structured memory by discovering and storing task-relevant rules, environmental patterns, and action-outcome relationships from test-time observations as natural language descriptions; 
and (2) an action-reasoning module that determines optimal actions through context-aware perception and strategic reasoning by dynamically retrieving and integrating knowledge from memory.
The action-reasoning module continuously updates its policy through proposed metacognitive test-time reinforcement learning, adapting as knowledge memory evolves.
We evaluate MCTR on 45 Atari games (33 seen, 12 unseen). MCTR demonstrates robust test-time adaptation, achieving 9/12 top-1 results on unseen games compared with baselines. Analyses through ablations, learning dynamics, and case studies reveal the complementary contributions of both components and show meta-reasoning evolving toward human-like adaptation strategies.
\end{abstract}    

\section{Introduction}
\label{sec:intro}

\begin{figure}
    \centering
    \includegraphics[width=\linewidth]{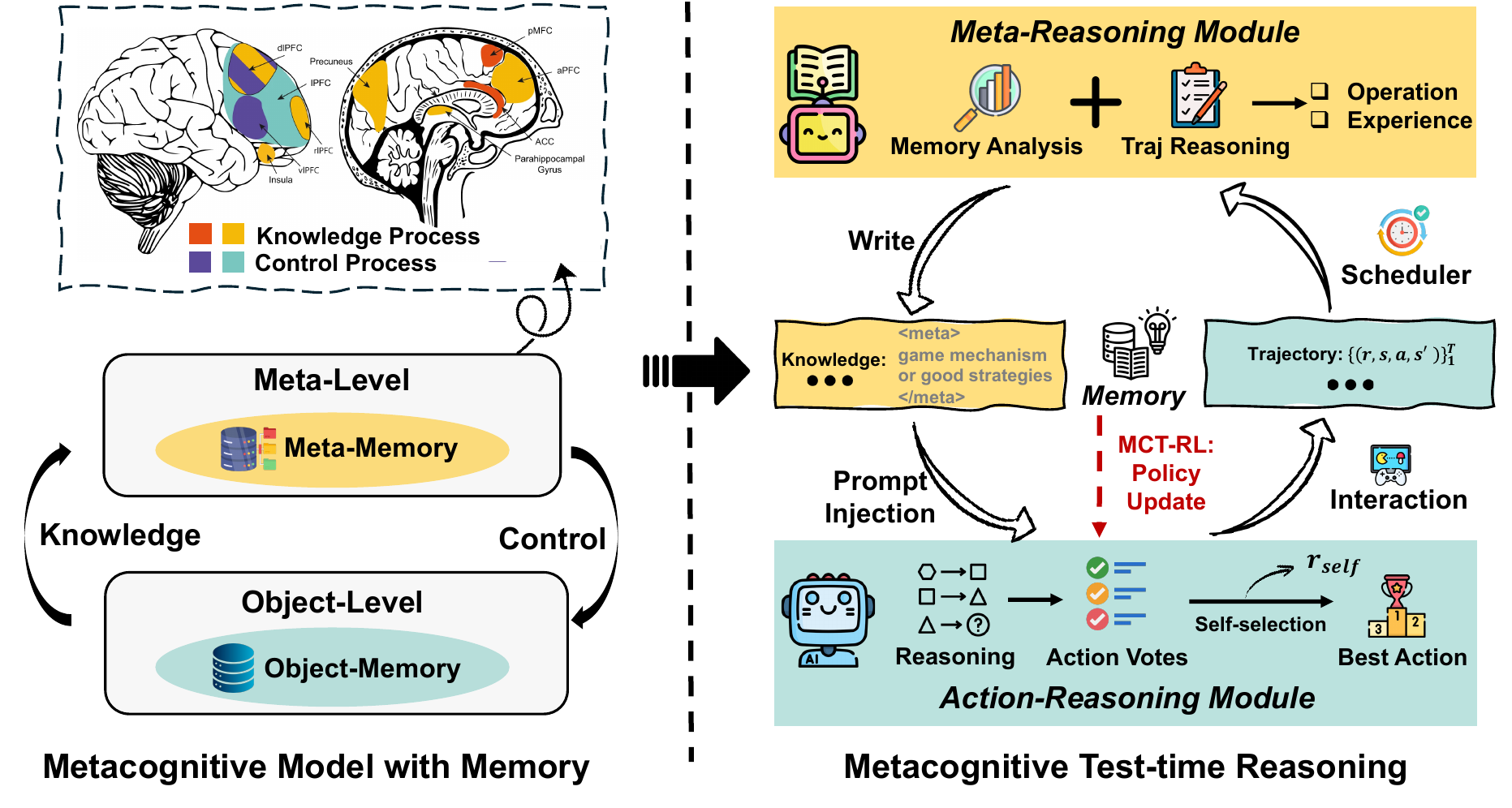}
    \caption{
    Comparison between a human metacognitive model and our MCTR framework.
    Left: A human metacognition model~\cite{nelson1990metamemory,Rivers2021Metacognition,Fleur2021} comprises meta-level and object-level processes with dedicated memory systems, enabling bidirectional knowledge and control flow. Brain regions associated with knowledge (orange) and control (green) processes are shown.
    Right: MCTR attains human-like adaptation to novel tasks without prior knowledge through two synergistic modules: a meta-reasoning module (orange) that discovers operational knowledge under dynamic scheduling, and an action-reasoning module (green) that leverages this knowledge for multi-step reasoning while adapting via test-time reinforcement learning.
    }
    \label{fig:intro}
    \vspace{-7mm}
\end{figure}

Large language models (LLMs) have made substantial progress towards general reasoning, driven by large-scale pre-training and neural scaling laws~\cite{gpt4o2024,gemini2024,deepseekv3,qwenvl2024}. Recent advances further improve reasoning through chain-of-thought prompting~\cite{wei2022chain}, tree-based search~\cite{yao2023tree}, and reinforced reasoning pipelines that use process reward models and trial-and-error self-play~\cite{xu2025large,deepseekr1,openai-o1,qwenreasoner2025}, giving rise to large reasoning models that scale both train time and test time computation~\cite{xu2025large}. Beyond scaling, LLMs display emergent abilities such as role playing~\cite{jiang2023roleplay} and analogical reasoning~\cite{webb2023emergent}, with applications in tool use, autonomous agents, and embodied decision-making. For adaptation at deployment, recent test-time approaches include inference-only methods such as in-context learning and prompt tuning~\cite{dong2024survey,shu2022tpt,liu2023incontext}, test-time training~\cite{sun2020test,akyurek2024surprising,ttrl2025}, and memory-based online reinforcement learning~\cite{zhou2025mementofinetuningllmagents} that refines policies using environmental feedback~\cite{snell2024scaling}.

Despite this progress, current vision–language reasoning models remain limited in fluid intelligence, the capacity to adapt to novel, complex tasks without prior knowledge~\cite{chollet2019measure,arcagi2025}.
Their abilities are largely crystallised, relying on accumulated knowledge and familiar patterns. Existing test time adaptation methods face further drawbacks. Test-time training~\cite{sun2020test,akyurek2024surprising} incurs non-trivial computational cost due to gradient updates during inference; prompt-based techniques~\cite{shu2022tpt} depend on augmentation or retrieval that transfers poorly to unfamiliar task structure; and recent test-time reinforcement learning (TTRL)~\cite{ttrl2025,zhang2024rest}, although able to discover strategies by trial and error without external rewards, remains limited in interpretability. Crucially, these methods do not exhibit fluid intelligence because they rely on implicit optimisation signals rather than constructing structured knowledge about task dynamics and reasoning metacognitively about when and how to apply strategies.

Humans facing novel tasks adapt from limited experience by observing successes, forming hypotheses, and refining strategies without extensive retraining. Cognitive science attributes this ability to metacognition~\cite{nelson1990metamemory,Rivers2021Metacognition,Fleur2021} (left panel, Fig.~\ref{fig:intro}), a two-level architecture with memory in which object-level processes execute tasks while meta-level processes monitor and control them in real time. Empirical studies show that effective learners revise reasoning strategies mid-task without external feedback~\cite{roebers2017metacognition,dinsmore2023metacognition,efklides2023metacognition}, achieving rapid adaptation with high sample efficiency~\cite{lake2017building}. This perspective motivates endowing vision-language models with analogous meta-level capabilities: to reflect on their own traces, evaluate effectiveness online, and adjust strategies from limited experience.

Motivated by this, we propose \textbf{MetaCognitive Test-Time Reasoning (MCTR)}, 
a framework that equips vision–language models with human-like adaptive capabilities through structured meta-reasoning and experience-driven learning during inference.
As shown in Fig.~\ref{fig:intro}, MCTR comprises two modules aligned with the metacognitive view. The meta-reasoning module performs retrospective analysis of accumulated trajectories, discovers operational patterns and strategies for new tasks, and stores them as natural language entries in a knowledge memory. 
A scheduler regulates meta-reasoning module invocation based on knowledge stability, prioritising frequent summarisation when knowledge is sparse and reducing overhead as the memory matures. 
The action-reasoning module conditions on this knowledge to conduct mul-step vision-language reasoning and updates its policy through metacognitive test-time reinforcement learning with self-consistency rewards, enabling online adaptation when external supervision is absent or feedback is weak.

We evaluate MCTR on 45 Atari games (33 seen, 12 held-out) to assess generalisation. MCTR achieves 9/12 top-1 results on unseen games, demonstrating robust test-time adaptation.
To understand what drives this improvement, we conduct three complementary analyses. Ablation studies confirm both metacognitive reasoning and MCT-RL are essential for adaptation. Learning dynamics analysis reveals MCTR balances reasoning stability through multi-path consensus with genuine policy adaptation, evidenced by increasing agreement among current reasoning traces while declining alignment with historical actions. Besides, qualitative analysis shows meta-reasoning evolves from exploratory hypotheses to concrete, goal-directed strategies that mirror human adaptation patterns.
Together, these results indicate that metacognitive test-time reasoning provides a promising route to flexible and robust decision-making in open-ended domains.

The main contributions of this paper are threefold:
\begin{itemize}
\item We introduce metacognitive test-time reasoning, a new paradigm that enables vision–language models to self-reflect and learn adaptively during inference.
\item We develop a complete MCTR pipeline, including (i) pre-deployment supervised reasoning fine-tuning, and (ii) test-time adaptation through metacognitive reinforcement learning with memory accumulation and self-consistency rewards.
\item We demonstrate state-of-the-art zero-shot adaptation to twelve unseen long-horizon Atari games, with ablations confirming the necessity of metacognitive components for human-like flexibility.
\end{itemize}

\section{Related Work}
\label{sec:related_work}

\textbf{Vision-Language Models and Test Time Adaptation.}
Vision-language models show strong reasoning via chain-of-thought prompting~\cite{wei2022chain}, tree-based search~\cite{yao2023tree,hao2023reasoning}, and process reward models~\cite{xu2025large,deepseekr1,openai-o1,qwenreasoner2025}, with applications in embodied decision-making~\cite{brohan2023rt,driess2023palm}, games~\cite{lee2023visla,du2023guiding}, and robotics~\cite{ahn2022can,huang2023voxposer}. 
Yet behaviour remains largely crystallised, drawing on pre-training patterns rather than adapting to novel structures. 
Test-time adaptation enables adjustment to new distributions without retraining~\cite{sun2020test,wang2021tent}. 
Prompt-based methods~\cite{shu2022tpt,liu2023incontext,dong2024survey} rely on in-context learning but struggle with unfamiliar structure; test-time training~\cite{sun2020test,akyurek2024surprising,gandelsman2022ttt} performs gradient updates at inference and is costly; test-time reinforcement learning~\cite{ttrl2025,zhang2024rest,snell2024scaling} exploits environmental feedback but requires extensive interaction. 
Alternatives use entropy minimisation~\cite{wang2021tent,niu2022efficient} or perplexity reduction~\cite{hu2025ttl}. 
Fluid intelligence, the ability to solve novel problems without prior knowledge, remains challenging~\cite{chollet2019measure}. ARC~\cite{arcagi2025,chollet2024arc} targets rule induction from few examples, with approaches spanning programme synthesis~\cite{acquaviva2022communicating}, neurosymbolic reasoning~\cite{hodel2023solving}, and large language models~\cite{greenblatt2024arc}. Unlike methods using statistical proxies or meta-training, our approach explicitly evaluates adaptation quality through metacognitive self-reflection, linking cognitive principles with practical test-time adaptation.

\noindent\textbf{Cognition-inspired Reasoning Models.}
Cognitive theories motivate architectures that integrate human-like processes. Kahneman’s dual process account~\cite{kahneman2011thinking} separates fast, intuitive judgements (System~1) from slow, deliberate reasoning (System~2), a distinction realised in recent systems~\cite{liu2025system12,yang2024llm2,he2024dpt,wei2025reasoning}. Examples include LLM2, which combines generation with process-based verification~\cite{yang2024llm2}, and SOFAI, which coordinates heuristic and deliberate routines for adaptability~\cite{sofai2025}. Hierarchical models extend this across levels of abstraction; HRM links high level planning with low level execution~\cite{sun2025hrm}. A complementary line studies metacognition, which monitors and controls one’s own inference~\cite{cox2005metacognition,johnson2022metacognition,nelson1990metamemory}. Classical architectures separate object level reasoning from meta level control~\cite{cox2005metacognition}, while recent work explores self reflection in language models~\cite{shinn2023reflexion,madaan2023selfrefine,ji2025metacognitive,park2024metacognition}. Many approaches require meta training~\cite{finn2017maml,nichol2018reptile} or focus on single turn refinement. We instantiate the dual level architecture of Nelson and Narens~\cite{nelson1990metamemory} for test time reasoning, enabling continuous metacognitive adaptation through structured monitoring and control without meta training.

\section{Preliminaries}

We briefly review two paradigms central to test-time adaptation: Test-Time Training (TTT) and Test-Time Reinforcement Learning (TTRL).

\noindent\textbf{Test-time Training.} 
TTT enables parametric models to adapt during inference through dynamic parameter updates~\cite{akyurek2024surprising}. For each test input (or batch) $x_{\text{test}}$, TTT constructs a temporary training dataset $\mathcal{D}_{\text{TTT}}$ and optimizes model parameters to minimize a loss function:
\begin{equation}
\theta_{\text{test}} = \arg\min_{\theta} \sum_{d_{\text{TTT}} \in \mathcal{D}_{\text{TTT}}} \mathcal{L}(\text{LM}(d_{\text{TTT}}; \theta)).
\end{equation}
The adapted parameters $\theta_{\text{test}}$ are then used for prediction. Previous work typically constructs $\mathcal{D}_{\text{TTT}}$ using unsupervised objectives applied to $x_{\text{test}}$ alone. In few-shot settings, we extend TTT by treating demonstration examples $\{(x_k, y_k)\}_{k=1}^K$ as supervised training instances. 

\noindent\textbf{Test-time Reinforcement Learning.} To address scenarios where labeled supervision is unavailable during deployment, TTRL~\cite{ttrl2025} reformulates test-time adaptation as a sequential decision problem. The model maintains a stochastic policy $\pi_\theta(y|x)$ that generates responses $y$ conditioned on input prompts $x$, where $\theta$ represents the adaptable parameters.
TTRL constructs self-supervised training signals through output self-consistency. For each test input $x$, the model samples a collection of $N$ candidate responses $\mathcal{Y} = \{y_1, y_2, \ldots, y_N\}$ from its current policy. These candidates are then aggregated—typically through majority voting or answer clustering—to produce a consensus prediction $\hat{y}$. The intuition is that this consensus captures the model's most confident prediction, which can serve as a reliable pseudo-label for parameter refinement.
The adaptation objective encourages the policy to assign higher probability to responses that align with the consensus:
\begin{equation}
\mathcal{L}_{\text{TTRL}}(\theta) = -\mathbb{E}_{y \sim \pi_{\theta}(\cdot | x)}[\text{sim}(y, \hat{y}) \cdot \log \pi_{\theta}(y | x)],
\end{equation}
where $\text{sim}(y, \hat{y})$ quantifies the similarity between generated output $y$ and consensus $\hat{y}$. Parameters are updated via gradient descent to minimize this loss, effectively steering the model toward more self-consistent predictions. This approach enables unsupervised test-time adaptation by exploiting the model's internal agreement structure rather than relying on external labels.
\section{Methodology}
\label{sec:method}
\begin{figure*}[h]
    \centering
    \includegraphics[width=0.9\linewidth]{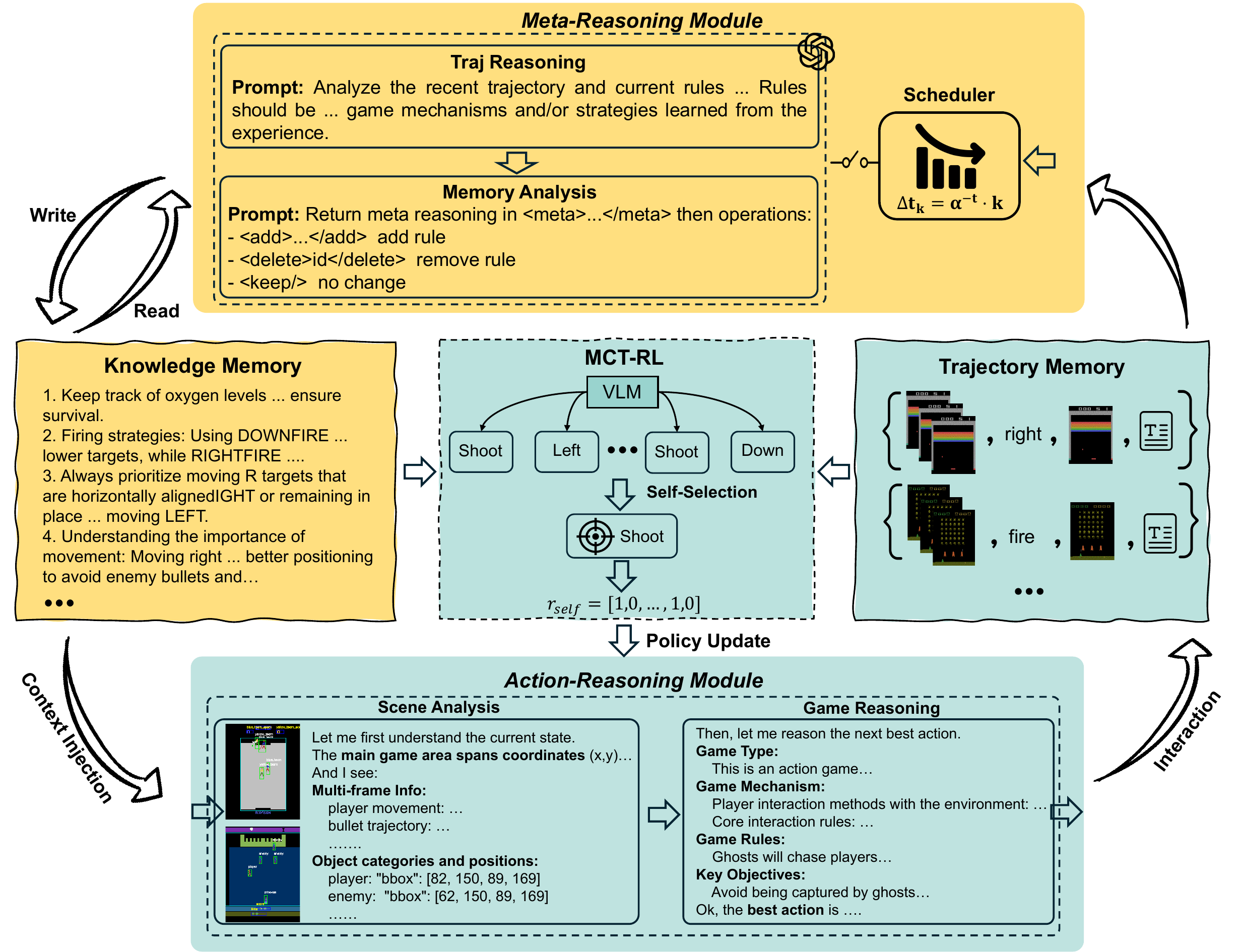}
    \caption{\textbf{Overview of Meta-Cognitive Test-Time Reasoning (MCTR) Framework.} The \textbf{meta-reasoning module} (yellow, meta-level process in human metacognitive model) performs retrospective analysis on accumulated trajectory memory and generates memory operations (\texttt{<add>}, \texttt{<delete>}, \texttt{<keep>}) to update knowledge memory. A scheduler dynamically adjusts meta-reasoning frequency to prioritize frequent analysis when knowledge is sparse and reduce invocations as memory matures. The \textbf{meta-reasoning module} (green, object-level process) handles real-time decision-making by injecting knowledge into context to guide multi-step VLM reasoning. The ARM continuously refines its policy at test time through reinforcement learning using self-supervised signals from action majority voting and real-time knowledge and trajectory memory.}
    \label{fig:method}
    \vspace{-5mm}
\end{figure*}


Drawing on the hierarchical structure of human metacognition, MCTR endows vision-language models with adaptive test-time reasoning through (1) a meta-reasoning module for reflective knowledge formation and (2) an meta-reasoning module for knowledge-guided decision-making.

As illustrated in Fig.~\ref{fig:method}, MCTR instantiates this dual-process architecture via two synergistic modules. 
Governed by an adaptive scheduler, meta-reasoning module performs memory analysis and trajectory reasoning to discover operational experiences, such as task mechanisms and strategies, stored as natural language descriptions. 
The scheduler prioritizes frequent analysis when knowledge is sparse and gradually reduces invocations as memory matures.
Complementarily, meta-reasoning module leverages these experiences both non-parametrically, through prompt-based context injection to guide reasoning, and parametrically, via metacognitive test-time reinforcement learning (MCT-RL) at every fixed timestep interval, to continually refine its policy as the knowledge base evolves. 
Crucially, MCT-RL enables the meta-reasoning module to optimize its reasoning and decision-making policy while efficiently reusing test-time data under no external supervision. 

The remainder of this section is organised as follows: Section~\ref{sec:method_meta} details the meta-reasoning module; Section~\ref{sec:method_action} describes the meta-reasoning module.

\subsection{Meta-Reasoning Module}
\label{sec:method_meta}

We now describe the meta-reasoning module, as shown at the top of Fig.\ref{fig:method}, which operates at the metacognitive level to discover task knowledge from interaction experience. Meta-reasoning module maintains an explicit knowledge memory $\mathcal{M}_t$ that guides both VLM reasoning and policy optimization (MCT-RL).

Rather than invoking meta-reasoning module at every timestep, an adaptive scheduler dynamically modulates meta-reasoning frequency. Let $k_t$ denote the interval (in timesteps) between consecutive meta-reasoning module activations. The scheduler amplify this interval exponentially:
\begin{equation}
k_{t+1} = \text{clip}\left( k_t / \gamma, k_{\min}, k_{\max} \right)
\end{equation}
Where $\gamma < 1$ controls the adaptation rate, and $\text{clip}(\cdot, k_{\min}, k_{\max})$ enforces interval bounds. This mechanism naturally prioritizes frequent meta-reasoning during early exploration when $\mathcal{M}_t$ is sparse and knowledge is rapidly evolving, while reducing computational overhead as the knowledge base stabilizes and matures.

Upon activation at timestep $t$, meta-reasoning module retrieves a trajectory segment $\tau_{[t-k:t]} = \{(s_i, a_i,r^{env}_i, r^{self}_i, s_{i+1})\}_{i=t-k}^{t}$ spanning the most recent $k$ timesteps. The VLM $f_{\phi}$ then performs sequential generation to analyze this experience and update the knowledge base $\mathcal{M}_t = \{e_1, e_2, \ldots, e_{N_t}\}$, which stores task understanding as a collection of structured rules. The memory is initialized empty ($\mathcal{M}_0 = \emptyset$) and evolves through iterative meta-reasoning cycles. The generation process factorizes into two stages:
\begin{equation}
\begin{aligned}
& p_{\phi}(\Omega_t, \mu_t \mid \tau_{[t-H:t]}, \mathcal{M}_{t-1}) \\
=& \, p_{\phi}(\mu_t \mid \tau_{[t-k:t]}, \mathcal{M}_{t-1}) \cdot p_{\phi}(\Omega_t \mid \mu_t, \tau_{[t-k:t]}, \mathcal{M}_{t-1}),
\end{aligned}
\end{equation}
where the process first generates a meta-analysis $\mu_t \sim p_{\phi}(\cdot \mid \tau_{[t-k:t]}, \mathcal{M}_{t-1})$ that articulates observed patterns, strategies, and insights from the trajectory, formatted within \texttt{<meta>...</meta>} tags. Conditioned on this analysis, the model then generates a sequence of memory operations $\Omega_t \sim p_{\phi}(\cdot \mid \mu_t, \tau_{[t-k:t]}, \mathcal{M}_{t-1})$, where each operation takes the form $\omega \in \{\texttt{<add>}(e), \texttt{<delete>}(\text{id}), \texttt{<keep>}\}$. These operations modify $\mathcal{M}_{t-1}$ to produce the updated knowledge base $\mathcal{M}_t$, which is subsequently provided to the meta-reasoning module for knowledge-conditioned decision-making and to MCT-RL for policy optimization.

\subsection{Action-Reasoning Module}
\label{sec:method_action}

As shown at the bottom of Fig.~\ref{fig:method}, the meta-reasoning module serves as the execution engine of our framework, responsible for real-time environment interaction and continuous policy refinement. It integrates metacognitive insights with visual perception to adapt to unknown environments. We model the test-time adaptation problem as an unknown Markov Decision Process (MDP) defined by the tuple $\mathcal{E} = \langle \mathcal{S}, \mathcal{A}, \mathcal{T}, r, \gamma, \rho_0 \rangle$, where $\mathcal{S}$ denotes the pixel-level state space, with each state composed of the three latest frames of the game; $\mathcal{A}$ is the discrete action space containing legal in-game movements; $\mathcal{T}$ represents the unknown transition dynamics; $r$ is the reward function, which may be sparse and difficult for credit assignment; $\gamma$ is the discount factor; and $\rho_0$ is the initial state distribution. The objective is to learn a policy $\pi_\theta: \mathcal{S} \times \mathcal{M} \rightarrow \Delta(\mathcal{A})$, parameterised by $\theta$, which conditions on both the observation history (state $s_t$) and the dynamic knowledge memory $\mathcal{M}$ to maximise the expected cumulative return:
\begin{equation}
J(\theta, \mathcal{M}) = \mathbb{E}_{\tau \sim \pi_\theta(\cdot|\mathcal{M})}\left[\sum_{t=0}^{T} \gamma^t r(s_t, a_t)\right],
\end{equation}
where $\tau = (s_0, a_0, r_0, \ldots)$ denotes an interaction trajectory.


The policy $\pi_\theta$ is implemented via a Vision-Language Model (VLM) $f_\theta$ that executes a two-stage autoregressive generation process. 
In the first stage, the VLM analyses the visual input $s_t$ to generate structured semantic tokens $z_t \sim \pi_\theta(\cdot | s_t)$. The representation $z_t = (pos_t, cat_t, dyn_t)$ encodes the game images, containing object coordinates ($pos_t$), object category ($cat_t$), and dynamic attributes ($dyn_t$) such as moving direction. In the second stage, the VLM conditions on the generated scene $z_t$ and the external knowledge $\mathcal{M}$ to produce the response. This response emerges sequentially: the model first generates a reasoning chain $r_t$ that explicates game understanding grounded in $\mathcal{M}$ (including mechanics and rules), followed by the final executable action $a_t$. 
The decision process is thus modelled as:
\begin{equation}
\pi_\theta(a_t | s_t, \mathcal{M}) = \sum_{z_t \in \mathcal{Z}} \pi_\theta(a_t | z_t, \mathcal{M}) \cdot \pi_\theta(z_t | s_t),
\end{equation}
where $\mathcal{Z}$ is the space of structured scene representations.
This formulation realises a knowledge-augmented policy where abstract game knowledge directly modulates object-level control as aligned with human metacognitive model in Fig.~\ref{fig:intro}.


\begin{table*}[!t]
\centering
\scriptsize
\caption{
Performance comparison on 33 seen Atari games and 12 unseen games. The upper block shows results on seen games used for the SFT phase, while the lower block  presents results on unseen games. The first four columns report pretrained VLMs without task-specific fine-tuning. The last columns show ablations removing: both components (w/o RL\&MR), meta-reasoning (w/o MR), and MCT-RL (w/o RL). 
\textbf{Bold} numbers indicate the best performance for each game, while \underline{underlined} numbers denote the best performance within the ablation studies. ``Top-1 Count'' rows report the number of games where each method achieves the highest score.
}
\label{tab:all_res}

\resizebox{\textwidth}{!}{
\begin{tabular}{ccccccccc}
\toprule
\multirow{2}{*}{\textbf{Game}} &
\multicolumn{4}{c}{\textbf{Pretrained Methods}} &
\multicolumn{4}{c}{\textbf{Finetuned Methods}} \\
\cmidrule(lr){2-5}\cmidrule(lr){6-9}
& \textbf{Gemini-2.5} & \textbf{GPT-4o} & \textbf{Qwen2.5-VL-7B} & \textbf{DeepSeek-VL2} &
\textbf{MCTR w/o RL\&MR} & \textbf{MCTR w/o MR} & \textbf{MCTR w/o RL} & \textbf{MCTR} \\
\toprule

\rowcolor{gray!10}
Alien & 0 & 30 & 0 & 490 & \textbf{10160} & / & / & / \\
\rowcolor{gray!10}
Amidar & 0 & 1 & 30 & 16 & \textbf{988} & / & / & / \\
\rowcolor{gray!10}
Assault & 42 & 42 & 189 & 21 & \textbf{399} & / & / & / \\
\rowcolor{gray!10}
Asterix & 850 & 500 & 550 & 350 & \textbf{6100} & / & / & / \\
\rowcolor{gray!10}
Atlantis & 0 & 0 & 0 & 4000 & \textbf{51600} & / & / & / \\
\rowcolor{gray!10}
BankHeist & 0 & 0 & 0 & 0 & \textbf{30} & / & / & / \\
\rowcolor{gray!10}
BeamRider & 0 & 0 & 176 & \textbf{528} & 176 & / & / & / \\
\rowcolor{gray!10}
Bowling & 12 & 0 & 12 & 3 & \textbf{86} & / & / & / \\
\rowcolor{gray!10}
Boxing & -14 & \textbf{-12} & -19 & -29 & -882 & / & / & / \\
\rowcolor{gray!10}
Breakout & 2 & 11 & 7 & 11 & \textbf{50} & / & / & / \\
\rowcolor{gray!10}
ChopperCommand & 300 & 0 & 800 & 1100 & \textbf{8700} & / & / & / \\
\rowcolor{gray!10}
DemonAttack & 0 & 0 & 60 & \textbf{295} & 40 & / & / & / \\
\rowcolor{gray!10}
DoubleDunk & \textbf{-2} & -12 & \textbf{-2} & -12 & -6 & / & / & / \\
\rowcolor{gray!10}
ElevatorAction & \textbf{0} & \textbf{0} & \textbf{0} & \textbf{0} & \textbf{0} & / & / & / \\
\rowcolor{gray!10}
FishingDerby & -54 & -54 & -54 & \textbf{-46} & -123 & / & / & / \\
\rowcolor{gray!10}
Gopher & 0 & 20 & 20 & 0 & \textbf{900} & / & / & / \\
\rowcolor{gray!10}
Hero & 150 & 0 & 0 & 0 & \textbf{28625} & / & / & / \\
\rowcolor{gray!10}
JourneyEscape & -20200 & -52700 & -78300 & \textbf{-11700} & -53800 & / & / & / \\
\rowcolor{gray!10}
Kangaroo & 0 & 0 & 0 & 0 & \textbf{2000} & / & / & / \\
\rowcolor{gray!10}
Krull & 40 & 280 & 0 & 0 & \textbf{2170} & / & / & / \\
\rowcolor{gray!10}
KungFuMaster & \textbf{0} & \textbf{0} & \textbf{0} & \textbf{0} & \textbf{0} & / & / & / \\
\rowcolor{gray!10}
Phoenix & 220 & \textbf{240} & 200 & 80 & \textbf{240} & / & / & / \\
\rowcolor{gray!10}
Pong & -27 & \textbf{-23} & -27 & -26 & -67 & / & / & / \\
\rowcolor{gray!10}
Qbert & 0 & 0 & 0 & 200 & \textbf{6850} & / & / & / \\
\rowcolor{gray!10}
Riverraid & 120 & 0 & 3430 & 960 & \textbf{8320} & / & / & / \\
\rowcolor{gray!10}
RoadRunner & 0 & 0 & 0 & 1400 & \textbf{42500} & / & / & / \\
\rowcolor{gray!10}
SpaceInvaders & 0 & 0 & \textbf{525} & 345 & 100 & / & / & / \\
\rowcolor{gray!10}
StarGunner & 300 & 0 & 0 & 300 & \textbf{1200} & / & / & / \\
\rowcolor{gray!10}
TimePilot & 400 & 400 & 400 & 600 & \textbf{2100} & / & / & / \\
\rowcolor{gray!10}
Tutankham & 0 & 0 & 1 & 0 & \textbf{2} & / & / & / \\
\rowcolor{gray!10}
UpNDown & 400 & 580 & 0 & 590 & \textbf{7290} & / & / & / \\
\rowcolor{gray!10}
WizardOfWor & 0 & 0 & 1000 & 200 & \textbf{4100} & / & / & / \\
\rowcolor{gray!10}
Zaxxon & \textbf{0} & \textbf{0} & \textbf{0} & \textbf{0} & \textbf{0} & / & / & / \\
\rowcolor{yellow!15}
\textbf{Top-1 Count (seen)} & 2/33 & 3/33 & 2/33 & 5/33 & \textbf{23/33} & / & / & / \\
\midrule

\rowcolor{pink!30}
AirRaid & \textbf{1125} & 150 & 0 & 175 & 125 & 200 & 75 & \underline{475} \\
\rowcolor{pink!30}
BattleZone & 2000 & 0 & 0 & 2000 & 5000 & 0 & 6000 & \underline{\textbf{12000}} \\
\rowcolor{pink!30}
Berzerk & 550 & 450 & 750 & 350 & 650 & 150 & 1000 & \underline{\textbf{1100}} \\
\rowcolor{pink!30}
Carnival & 1280 & 960 & 20 & 740 & 600 & 500 & 360 & \underline{\textbf{2660}} \\
\rowcolor{pink!30}
CrazyClimber & 0 & 0 & 0 & 0 & 1100 & 1000 & 1800 & \underline{\textbf{5600}} \\
\rowcolor{pink!30}
Freeway & 0 & 5 & 0 & 1 & 17 & \underline{\textbf{23}} & 22 & \underline{\textbf{23}} \\
\rowcolor{pink!30}
Frostbite & 0 & 0 & 0 & 140 & 310 & 240 & 430 & \underline{\textbf{650}} \\
\rowcolor{pink!30}
IceHockey & \textbf{-2} & \textbf{-2} & \textbf{-2} & -6 & -11 & -16 & \underline{-8} & -12 \\
\rowcolor{pink!30}
NameThisGame & 0 & 0 & 590 & 740 & 1720 & 70 & 1630 & \underline{\textbf{2170}} \\
\rowcolor{pink!30}
Robotank & 1 & 0 & 2 & 4 & \underline{\textbf{5}} & \underline{\textbf{5}} & 3 & \underline{\textbf{5}} \\
\rowcolor{pink!30}
Seaquest & 160 & 0 & 0 & 0 & 20 & 40 & 80 & \underline{\textbf{660}} \\
\rowcolor{pink!30}
Skiing $\downarrow$ & -6663 & -6663 & -6663 & -6656 & -20074 & \underline{\textbf{-30057}} & -29452 & -20334 \\
\rowcolor{yellow!15}
\textbf{Top-1 Count (unseen)} & 2/12 & 1/12 & 1/12 & 0/12 & 1/12 & 3/12 & 0/12 & \textbf{9/12} \\
\bottomrule
\end{tabular}
}
\vspace{-3mm}
\end{table*}

To equip the meta-reasoning module with robust initial capabilities, we perform SFT on our specialised reasoning dataset. We augment raw gameplay trajectories with high-quality natural language rationales generated by a teacher model. By training on these traces, comprising a reflective \texttt{<think>} analysis and an \texttt{<answer>} segment, the meta-reasoning module is initialised with strong priors for scene understanding and logical deduction before entering the test-time phase. Once the model learns the reasoning pattern, we optimise the model at the test-time both non-parametrically by injecting the memory into the model context, and parametrically by reinforcement learning paradigm MCT-RL.

In MCT-RL, the meta-reasoning module follows a periodic learning process inspired by human metacognition, interleaving active knowledge acquisition with online optimization.
This reflection further enhances the action-reasoning module's ability to effectively leverage acquired recent knowledge for problem-solving. 
We apply MCT-RL at every fixed interval of $T$ steps to explore promising reasoning paths by reusing past states $\{s_t\}_{t=|\tau|-T}^{|\tau|}$ stored in the trajectory memory $\mathcal{D}$. During test-time interaction, all experiences are continuously recorded in real-time into $\mathcal{D}$, forming an accumulated buffer $\mathcal{D} = {(s_t, a_t, r^{env}, r^{self}, s_{t+1})}_{t=1}^{|\tau|}$. This buffer jointly supports knowledge extraction in the meta-reasoning module and policy optimization in MCT-RL, enabling the agent to iteratively refine its reasoning and decision-making based on recent experiences.

To address the challenge of sparse environmental rewards in long-horizon tasks, we replace traditional sparse environmental rewards with an internal, reasoning-based signal derived through majority voting~\cite{ttrl2025}. For each past state $s_t$ at timestep $t$, the model samples $K$ candidates with explicit reasoning rationales:
\begin{equation}
\{a_i\}_{i=1}^K \sim \prod_{i=1}^K p_{\theta}(\cdot \mid s_t, \mathcal{M}),
\end{equation}
Candidates are evaluated via majority voting to extract the golden action:
$
a_t^* = \arg\max_{a \in \mathcal{A}} \sum_{i=1}^{K} \mathbb{I}(a_i = a),
$
Each candidate is assigned by an indicator reward function  comparing candidate with the golden action:
\[
r_t(s_t, a) = \mathbb{I}(a = a_t^*) =
\begin{cases}
1, & \text{if } a = a_t^*, \\
0, & \text{otherwise.}
\end{cases}
\]
The policy is updated to maximise consistency with this consensus using GRPO~\cite{shao2024deepseekmath} algorithm, where the final objective can be written as:
\begin{equation}
\begin{aligned}
\mathcal{J}_{\mathrm{GRPO}}(\theta)
&= \mathbb{E}_{s_t \sim S_\mathcal{D}[-T:],\, \{a_i\}_{i=1}^K \sim \pi_{\theta_{\text{old}}}(\cdot \mid s_t)}
\Bigg[
\frac{1}{K} \sum_{i=1}^K 
\frac{1}{|a_i|} 
\sum_{t=1}^{|a_i|} \\
&\hspace{1.2em} \min\Big(
w_{i,t}(\theta)\,\widehat{A}_{i,t},\,
\operatorname{clip}\big(
w_{i,t}(\theta)
\big)\widehat{A}_{i,t}
\Big)
\Bigg],
\end{aligned}
\end{equation}
where the importance ratio $w_{i, t}(\theta)$ is defined as:
\begin{equation}
w_{i, t}(\theta) =
\frac{\pi_\theta\left(a_{i, t} \mid s_t\right)}
{\pi_{\theta_{\text{old}}}\left(a_{i, t} \mid s_t\right)},
\end{equation}
and the advantange $\widehat{A}_{i, t}$ for action $a_{i,t}$ is defined as:
\begin{equation}
\widehat{A}_{i, t} =
\frac{r_t\left(s_t, a_{i,t}\right) - 
\operatorname{mean}\left(\{r_t(s_t, a_{i,t})\}_{i=1}^G\right)}
{\operatorname{std}\left(\{r_t(s_t,  a_{i,t})\}_{i=1}^G\right)}.
\end{equation}



\section{Experiments}
\label{sec:exp}

In this section, we conduct comprehensive experiments to evaluate MCTR's effectiveness in test-time adaptation to novel environments. Our evaluation focuses on the Atari game benchmark, comprising 45 diverse games that challenge decision-making across varying strategic requirements and visual complexity. 
Specifically, Section~\ref{sec:exp_setting} introduces the experimental setting, Section~\ref{sec:results} presents the main results, and Section~\ref{sec:exp_case} provides a case study.

\begin{figure*}[t]
\centering
\begin{minipage}[t]{0.4\linewidth}
    \vspace{0pt}
    \centering
    \small
    \setlength{\tabcolsep}{4pt}
    \captionof{table}{\textbf{Ablation study on adaptive interval scheduling.} Comparison of different initial intervals (\(k\)) and growth rates (\(\lambda\)) across a subset of Atari games. Bold values indicate best performance per game.}

    \label{tab:ablation_k_lambda}
    \vspace{2mm} 
    \resizebox{\linewidth}{!}{%
    \begin{tabular}{ccrrrrr}
    \toprule
    \textbf{$k$} & \textbf{$\lambda$} & \textbf{IceHockey} & \textbf{BattleZone} & \textbf{AirRaid} & \textbf{Frostbite} & \textbf{Carnival} \\
    \midrule
    3  & 1.00  & -13 & 2000  & 300 & 310 & \underline{540} \\
    9  & 1.00  & -16 & 4000  & 100 & 370 & 360 \\
    15 & 1.00  & \underline{-12} & 2000  & 0   & 270 & 420 \\
    20 & 1.00  & \underline{-12} & \underline{6000} & \underline{325} & \underline{440} & \underline{440} \\
    15 & 1.18  & \textbf{-8} & \underline{6000} & 0   & 370 & 300 \\
    \rowcolor{yellow!12}
    3  & 0.85  & \underline{-12} & \textbf{12000} & \textbf{475} & \textbf{650} & \textbf{2660} \\
    \bottomrule
    \end{tabular}%
    }
\end{minipage}%
\hfill
\begin{minipage}[t]{0.57\linewidth}
    \vspace{0pt}
    \centering
    \caption{\textbf{MCT-RL dynamics analysis.} \textbf{Left:} Majority voting ratio increases as the agent progressively favors higher-quality, self-consistent actions. \textbf{Right:} Agreement ratio with historical trajectory memory declines, indicating active policy revision with newly acquired environmental knowledge.}
    \vspace{-1mm} 
    \begin{subfigure}[t]{0.48\linewidth}
        \centering
        \includegraphics[width=\linewidth]{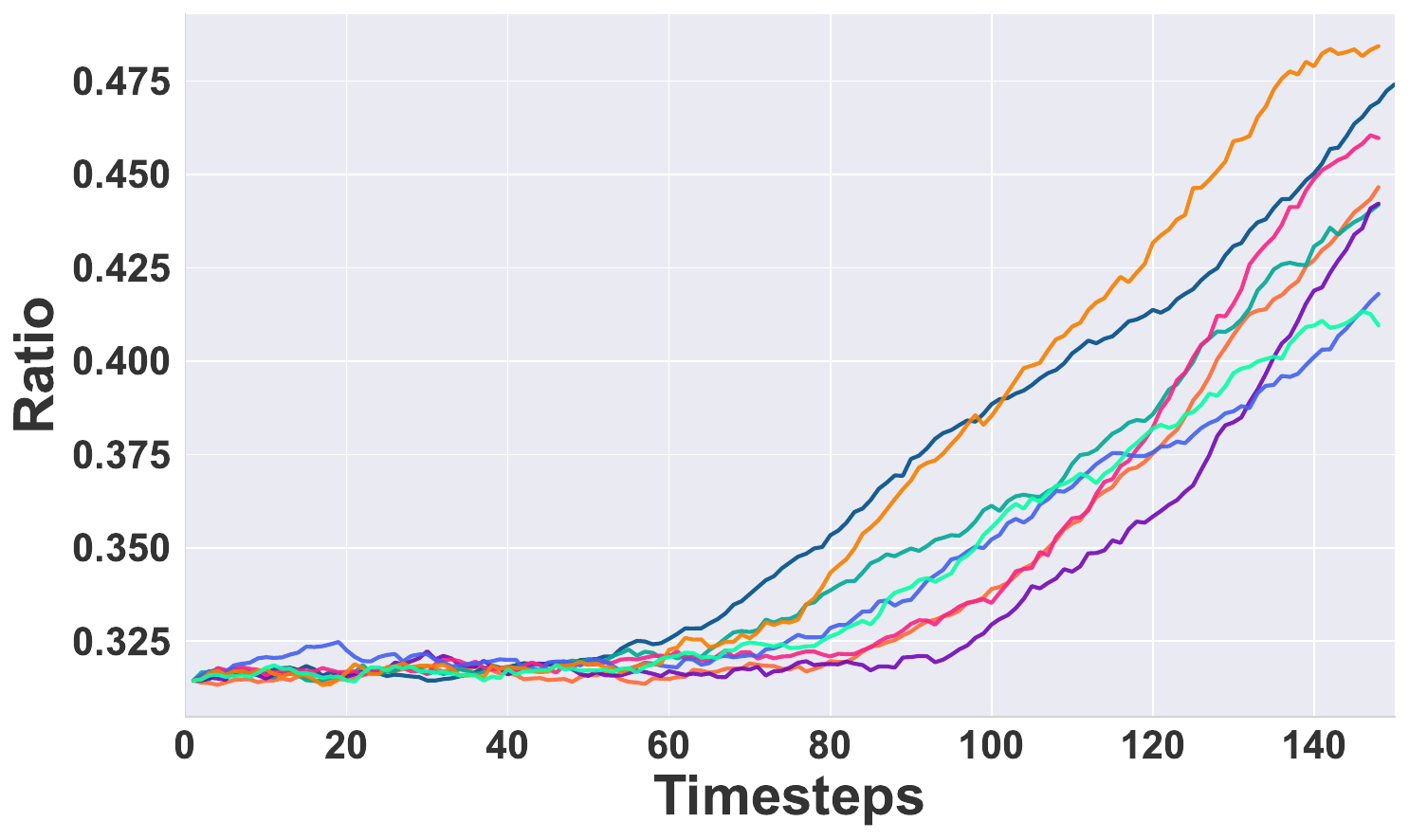}
        \label{fig:ttrl_majority_voting_reward}
    \end{subfigure}%
    \begin{subfigure}[t]{0.48\linewidth}
        \centering
        \includegraphics[width=\linewidth]{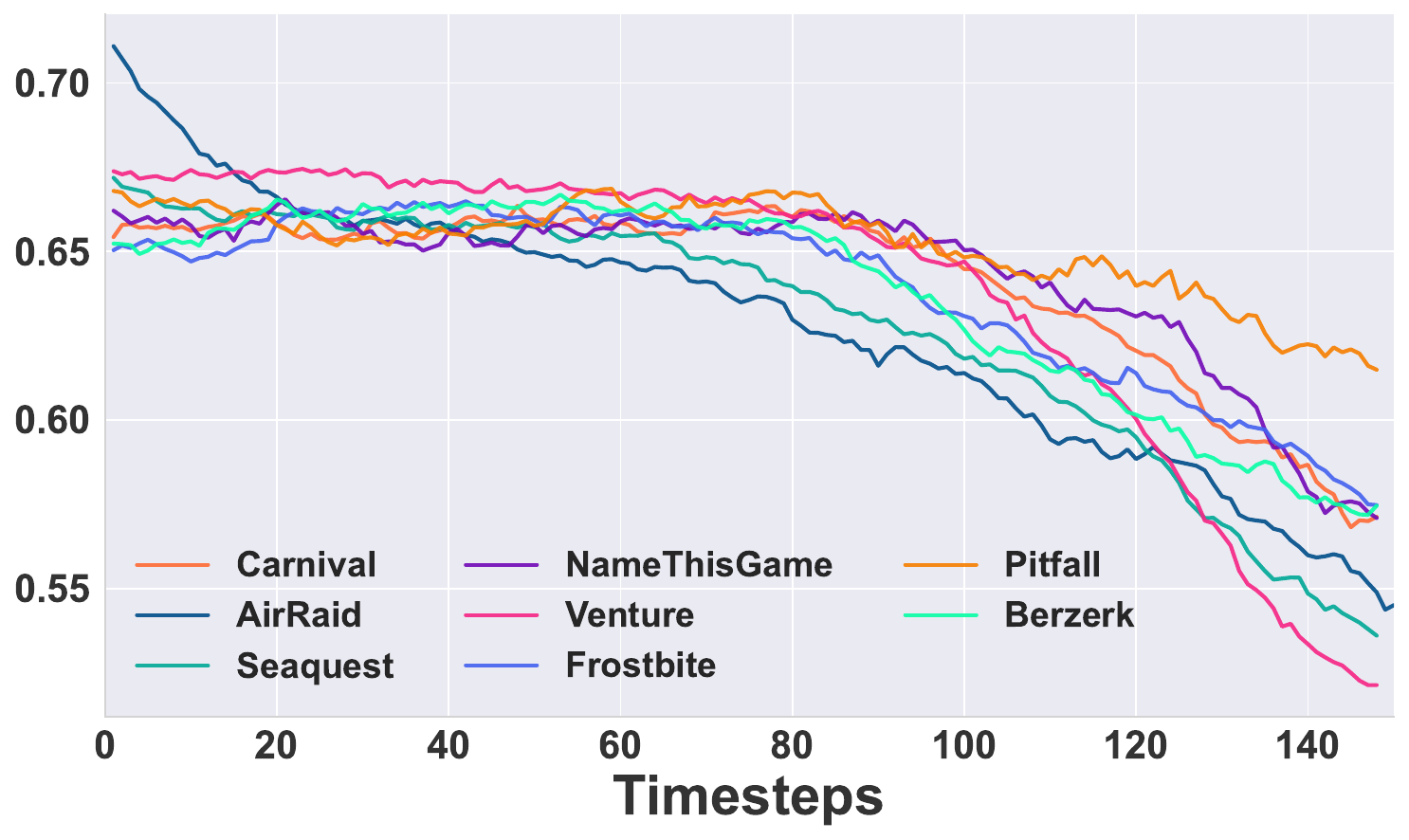}
        \label{fig:ttrl_reward_accuracy}
    \end{subfigure}
    \label{fig:ttrl_reward_analysis}
\end{minipage}
\vspace{-10mm}
\end{figure*}

\vspace{-2mm}
\subsection{Experimental Setting}
\label{sec:exp_setting}
The MCTR begins with supervised fine-tuning of \texttt{Qwen 2.5-7B-VL} as the base model. We train on 33 Atari games, collecting 10{,}000 samples per title using \texttt{DQN\_modern} \cite{gogianu2022agents} policies in the \texttt{NoFrameskip-v4} environment. Data collection operates on four stacked frames with $\varepsilon$-greedy or Boltzmann exploration, storing at each timestep a three-frame state representation, the executed action, the received reward, and a binary emulator snapshot. To generate language supervision for these trajectories, we first derive visual grounding through game-specific OpenCV settings that specify play regions, colour masks, and templates. These settings enable automatic extraction of bounding boxes for key entities within each frame. We then prompt Gemini 2.0 Flash with propositions parsed from these visual groundings to obtain stepwise reasoning traces. Each trace comprises a reflective \texttt{<think>} analysis followed by an \texttt{<answer>} that restates the action, ensuring alignment with the competent control demonstrated by the open-source DQN policies. The resulting SFT samples integrate three frames, the DQN action, extracted bounding boxes, and Gemini-generated reasoning. Training proceeds for six epochs across all games. Further details on dataset construction are provided in supplementary material.

During test-time adaptation, the meta-reasoning interval $k$ is initialised to 3 and dynamically adjusted according to $k \leftarrow k / 0.85$ after each meta-reasoning cycle, bounded within $[2, 15]$ to balance computational cost with adaptation responsiveness. The memory bank maintains a maximum capacity of 20 entries to store recent experiences. Following this initialization, the model undergoes MCT-RL every 100 interaction steps and each MCT-RL stage runs for five epochs with a group size of 8.

To enable efficient parameter updates during this adaptation phase, we employ Low-Rank Adaptation (LoRA) with rank-64 adapters and scaling factor $\alpha=32$ applied to all linear layers except the vision modules, which remain frozen to preserve visual understanding. This configuration maintains the base model's reasoning and perception capabilities whilst enabling lightweight fine-tuning with substantially fewer trainable parameters than full model updates. The prompting templates used for reasoning and reflection are provided in the supplementary material.

\subsection{Main Results}
\label{sec:results}

\textbf{Performance Comparison in Atari Games.}
Table~\ref{tab:all_res} presents comprehensive evaluation results across 45 Atari games, divided into 33 seen games used during supervised fine-tuning (upper block, gray) and 12 unseen games for generalisation testing (lower block, pink). We compare pretrained vision-language models, our supervised fine-tuning baseline, and ablation variants of our MCTR approach.

The SFT (MCTR w/o RL\&MR) baseline achieves the highest overall performance on seen tasks (23 out of 33 top-1 scores), demonstrating that supervised reasoning-augmented fine-tuning effectively aligns the model with expert policies, while general pretrained VLMs show limited competence with near-zero or negative rewards. However, this strong seen performance does not transfer to novel environments, the SFT baseline achieves only 1 out of 12 top-1 scores on unseen games. In contrast, our proposed MCTR achieves the best generalization on unseen unseen games, securing 9 out of 12 top-1 results. MCTR interleaves reflection-based meta-reasoning with MCT-RL, enabling both efficient adaptation and exploration in new environments. The performance gains are particularly pronounced in complex games: on BattleZone, MCTR achieves 12000 compared to 5000 for the SFT baseline; on CrazyClimber, it reaches 5600 versus 1100; and on Carnival, it attains 2660 versus 600. Notably, pretrained VLMs continue to struggle on unseen games (0 top-1 scores for the best baseline DeepSeek-VL2), highlighting the critical importance of test-time adaptation.

\noindent\textbf{Ablation Study.} As shown in Table~\ref{tab:all_res}, the last four columns reveal the contribution of each component. Removing both test-time RL and meta-reasoning (MCTR w/o RL\&MR) yields strong seen performance (23/33) but poor unseen generalization (1/12), indicating supervised fine-tuning alone cannot transfer to novel environments. Adding test-time RL alone (MCTR w/o MR) improves unseen performance to 3/12 top-1 scores, while adding meta-reasoning alone (MCTR w/o RL) achieves similar results but excels in different games, as MR favours strategic adjustment (e.g., BattleZone: 6000) while MCT-RL enables policy optimisation (e.g., Freeway: 23). The complete MCTR system achieves 9/12 top-1 scores, demonstrating strong synergy: meta-cognitive reflection guides exploration during test-time RL, as evidenced by substantial improvements on BattleZone (12000 vs. 6000) and CrazyClimber (5600 vs. 1800). This integration improves average unseen performance by 275\% over the SFT baseline.

\noindent\textbf{Effect of Adaptive Meta-Reasoning Scheduling.}
To verify the effect of different adaptive interval scheduling strategies, we conducted ablation studies across a subset of Atari games, as shown in Table~\ref{tab:ablation_k_lambda}. The adaptive exponential schedule (\(k{=}3\), \(\lambda{=}0.85\)) consistently outperforms both fixed-frequency baselines (\(\lambda{=}1.0\)) and slower growth variants (\(\lambda{=}1.18\)) across all tested environments, achieving particularly strong results on BattleZone (12000) and Carnival (2660).
Excessively frequent meta-reasoning intervals (e.g., \(k{=}3\), \(\lambda{=}1.0\)) demonstrate degraded performance as insufficient experience accumulates between reflections, introducing instability without generating new strategic knowledge. Conversely, large fixed intervals (e.g., \(k{=}20\)) provide inadequate reflective guidance during early adaptation when strategic reassessment is most critical. The exponential schedule addresses both limitations: it enables dense reflection during the exploration phase when information gain is maximal, then progressively reduces frequency as the policy converges, yielding efficient test-time adaptation.

\noindent\textbf{MCT-RL Learning Dynamics Analysis.} Fig.~\ref{fig:ttrl_reward_analysis} shows how MCT-RL enhances adaptive decision-making capabilities during test-time adaptation. 
The left panel illustrates the evolution of majority voting reward, showing that the framework progressively favors reasoning traces that produce higher-quality actions with stronger internal agreement. 
Conversely, the right panel reveals declining agreement between action regeneration in the MCT-RL phase-informed by newly acquired knowledge and historically selected actions stored in trajectory memory that were generated using outdated environmental understanding. 
This decreasing alignment signals active policy revision as the agent continuously refines its decision-making strategy based on accumulated experience.
Together, these dynamics show that MCT-RL achieves dual objectives: it maintains reasoning stability through multi-path consensus while simultaneously enabling genuine policy adaptation in response to evolving environmental knowledge.

\begin{figure}
    \centering
    \includegraphics[width=\linewidth]{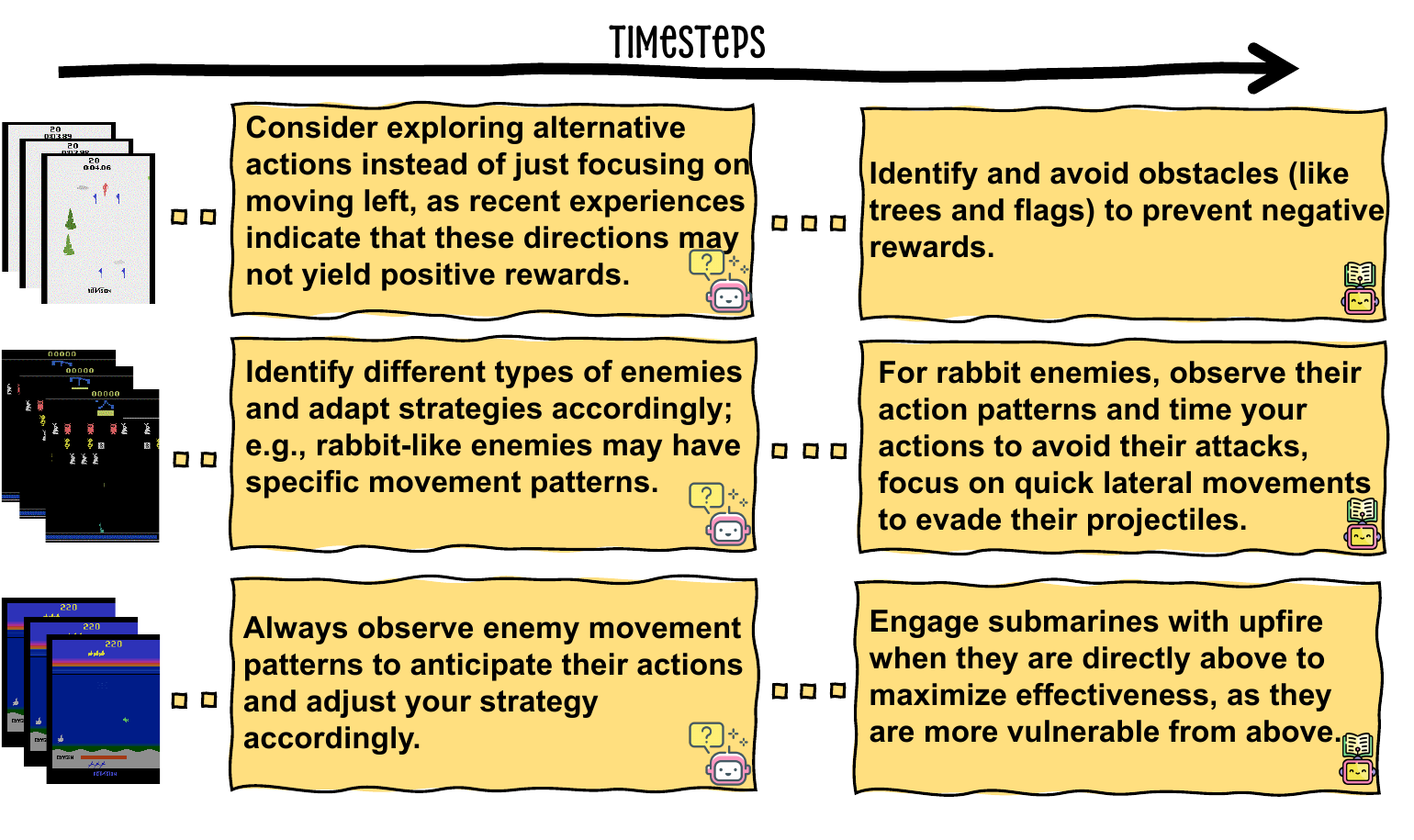}
    \vspace{-5mm}
    \caption{
    Case study illustrating the temporal evolution of meta-reasoning knowledge across three unseen games. Early reflections exhibit exploratory metacognition focused on hypothesis generation, while later reasoning crystallises into goal-directed, procedural strategies grounded in learnt environmental dynamics.
    }
    \label{fig:case-study}
    \vspace{-5mm}
\end{figure}

\subsection{Case Study}
\label{sec:exp_case}

Fig.~\ref{fig:case-study} presents a temporal analysis of meta-reasoning knowledge evolution across three unseen games, revealing a systematic progression from metacognitive exploration to situated expertise. 
The cases capture a typical transition in reasoning abstraction that parallels human skill acquisition.

Early-stage reasoning (left column) exhibits characteristic features of metacognitive monitoring under uncertainty. 
The module generates high-level epistemic directives, ``identify different types of enemies'' and ``consider exploring alternative actions'', that function as learning objectives rather than action policies.
This phase demonstrates active hypothesis generation, where the agent constructs a problem space by deliberately probing environmental contingencies. 
Notably, these early reflections emphasise what to learn rather than how to act, indicating metacognitive awareness of knowledge gaps.

Mid-to-late stage reasoning (right column) reveals a qualitative shift toward procedural expertise and situated decision-making. Abstract principles crystallize into concrete, context-specific strategies: precise timing patterns for rabbit enemies, spatial positioning tactics for submarine engagement, and obstacle avoidance heuristics. This progression reflects the compilation of declarative knowledge into procedural schemas, as knowledge becomes increasingly grounded in sensorimotor contingencies and reward structures specific to each game's dynamics.

This epistemic trajectory (from exploratory metacognition to goal-directed proceduralization) mirrors the cognitive transition observed in human skill acquisition, where initial controlled processing gives way to automated expertise. The meta-reasoning module thus exhibits genuine adaptive intelligence: it not only learns what patterns exist but dynamically regulates how to learn them, embodying a test-time instantiation of metacognitive model.

\section{Conclusion}
We presented Meta-Cognitive Test-Time Reasoning, a framework that equips vision-language models with human-like adaptive capabilities through structured meta-reasoning and experience-driven learning during inference. Drawing on human metacognition, MCTR implements a two-level architecture where a meta-reasoning module discovers and consolidates strategic knowledge from experience trajectories, while an action-reasoning module applies this knowledge for task execution and refines its policy online through our proposed metacognitive test-time reinforcement learning.
MCTR exhibits strong test-time generalization on the Atari benchmark, securing 9 out of 12 top-1 scores on unseen games. Ablation studies confirm that meta-reasoning and test-time learning provide complementary adaptation mechanisms: the former extracts transferable strategic patterns while the latter enables policy refinement through environmental feedback. Learning dynamics analysis demonstrates that MCT-RL balances reasoning stability through multi-path consensus with genuine policy adaptation, as evidenced by increasing internal agreement among current reasoning traces and declining alignment with historical actions. Qualitative analysis of reasoning traces reveals that strategies evolve from exploratory trial-and-error to structured, goal-directed patterns that mirror human adaptation to unfamiliar tasks.

{
    \small
    \bibliographystyle{ieeenat_fullname}
    \bibliography{main}
}
\clearpage
\setcounter{page}{1}
\maketitlesupplementary






\lstdefinestyle{prompt}{
  basicstyle=\ttfamily\small,
  breaklines=true,
  breakautoindent=false,
  breakindent=0pt,
  columns=fullflexible,
  xleftmargin=0pt,
  frame=single,
  framerule=0.4pt,
  rulecolor=\color{black!30},
  backgroundcolor=\color{black!3},
}

\setlength{\parskip}{6pt}
\setlength{\parindent}{0pt}




\section{Supervised Fine-tuning}
\label{sec:rationale}

We build our supervised fine-tuning pipeline  on top of the R1-V framework \citep{chen2025r1v}, with training details specified in the following sections.

\subsection{Reasoning Dataset Creation}

\paragraph{Data Collection.}
Existing Atari corpora rarely align step-level vision, text, and action or preserve restorable context, so we build a dataset that does. It provides images, parser summaries, and action-affirming rationales for studies of perception-to-decision reasoning and test-time adaptation. We release a stable schema, fixed splits, loaders, and documentation, and supply hashes where snapshots are restricted, enabling reproducible training and fair validation.

\begin{figure}[t]
\centering
\begin{subfigure}{0.24\linewidth}
    \centering
    \includegraphics[width=\linewidth]{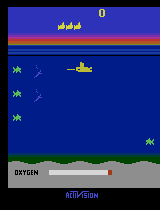}
    \caption{}
    \label{fig:suppl-70}
\end{subfigure}
\hfill
\begin{subfigure}{0.24\linewidth}
    \centering
    \includegraphics[width=\linewidth]{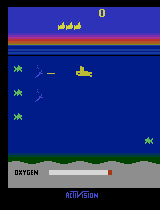}
    \caption{}
    \label{fig:suppl-71}
\end{subfigure}
\hfill
\begin{subfigure}{0.24\linewidth}
    \centering
    \includegraphics[width=\linewidth]{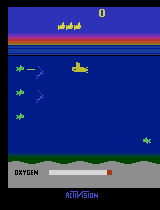}
    \caption{}
    \label{fig:suppl-72}
\end{subfigure}

\caption{Example of an observation which contains three images, from Atari game Seaquest.}
\label{fig:suppl-three-examples}
\end{figure}

We collect trajectories in Atari environments configured as \texttt{NoFrameskip-v4}. A \texttt{DQN\_modern} policy operates on stacks of three consecutive frames (Figure~\ref{fig:suppl-three-examples}) and selects discrete actions under either $\varepsilon$-greedy or Boltzmann exploration with a small default exploration rate. At the beginning of each episode the environment is reset. At every step we render the RGB frame, update a buffer containing the three most recent frames, and stack observations as a channel-first tensor for policy input.

We log the action identifier and its human-readable label, the immediate reward, and the running return. To support exact restoration during later analysis we archive a binary snapshot of the emulator runtime state. We maintain per-episode identifiers, step counters, and throughput statistics, and write summary files reporting average episode length and return. After the interaction budget is exhausted, we assemble the per-step records into Hugging Face \texttt{DatasetDict} objects with explicit train/test splits, cast image columns to the native image type, and save them to disk. The same rollout procedure builds both the supervised fine-tuning corpus and the reinforcement learning corpus; only the choice of games, interaction budget, and split assignment differ. Each game is accompanied by a version-controlled OpenCV configuration specifying the legal play region, color masks, object templates, thresholds, morphological kernels, and fixed UI counters for downstream parsing.

\paragraph{Language Reasoning with Gemini.}
For each recorded step we load the game-specific OpenCV configuration and parse the most recent frame within the legal region using template matching. Detected elements are stored and converted into concise natural language propositions describing class, location, and value when applicable. A short-term buffer retains the three most recent descriptions to provide immediate temporal context.

Using these descriptions together with the action label from DQN, we compose a prompt with fixed system and user instructions and query Gemini-2.0-Flash to obtain a step-level reasoning trace. The model returns a structured output comprising a reflective \texttt{<think>} segment and an \texttt{<answer>} segment that restates the chosen action, thereby pairing perception, interpretation, and decision as a single supervision signal. The complete prompt for Gemini is as below:

\begin{center}
\begin{lstlisting}[style=prompt]
You are shown three consecutive video frames from an Atari-style game. 
An oracle has already chosen its next move: {action}.

Your task is to adopt the oracle's perspective and articulate the reasoning behind this decision. **Do not** mention or describe the action itself in your output.

Begin by analyzing the visual understanding of each frame. For each timestep, you will be given a description of detected objects and their bounding boxes in the format: (x_top_left, y_top_left, x_bottom_right, y_bottom_right). All coordinates use a top-left origin, with the frame size being 210 pixels in height and 160 pixels in width.

- Frame 1 (timestep -2):  
{understanding_buffer[0]}

- Frame 2 (timestep -1):  
{understanding_buffer[1]}

- Frame 3 (current timestep):  
{understanding_buffer[2]}

Next, write a concise thought process-no more than 10 sentences-that logically explains why the oracle made this decision.
Your reasoning should be natural and grounded in the visual context, and you could reason it from following aspects but not limited to:

- **Game Genre Inference**: Based solely on visual cues, infer whether the game is action, shooting, strategy, or another genre.
- **Gameplay Mechanics**: Describe how the agent appears to interact with the environment (e.g., movement, avoidance, targeting) and how feedback or rewards might be structured.
- **Key Observations**: Point out any significant objects, obstacles, goals, or patterns of movement across the frames that likely influenced the decision.
- **Goal Analysis**: Explain what the agent seems to be trying to achieve immediately and over the long term.
- **Decision Justification**: Summarize why the inferred action is appropriate given the current game context-without stating or implying the action.

Avoid referencing the action name, any game titles, franchises, or platforms. Focus entirely on visual reasoning and inferred gameplay logic.

Please directly output your reasoning without any additional text or formatting.

\end{lstlisting}
\end{center}

We append these annotations to the rollout records and package them as Hugging Face datasets with train/test splits. Each sample contains the image sequence, actions, rewards, episode metadata, the restorable emulator state, the textual state descriptions, and the associated reasoning. The same parsing, prompting, and storage settings are applied in both the supervised fine-tuning and reinforcement learning stages so that data semantics and annotation protocols remain consistent.

\subsection{Example Data}

System prompt:

\begin{center}
\begin{lstlisting}[style=prompt]
You are a general-purpose game agent focused on scene understanding, key object recognition, and intelligent action decision-making based on continuous game frames. Please analyze and reason based on the following dimensions:

Scene Analysis: First, Extract dynamic information from the current frame and several previous frames to analyze the following:
- Object Categories: Identify entities such as the agent/player, enemies, obstacles, bullets, items, etc.
- Spatial Positions: For each object, use the bounding box format (x_top_left, y_top_left, x_bottom_right, y_bottom_right). The coordinate origin (0,0) is at the top-left of the frame.

Game Reasoning: Next, Based on the overall game context and the current scene, infer the most reasonable next atomic action. Your reasoning should integrate both static game rules and dynamic scene information, and potentially cover the following aspects:
- Game Genre Inference: Based solely on visual cues, infer whether the game is action, shooting, strategy, or another genre.
- Gameplay Mechanics: Describe how the agent appears to interact with the environment (e.g., movement, avoidance, targeting) and how feedback or rewards might be structured.
- Key Observations: Point out any significant objects, obstacles, goals, or patterns of movement across the frames that likely influenced the decision.
- Goal Analysis: Explain what the agent seems to be trying to achieve immediately and over the long term.
- Decision Justification: Summarize why the inferred action is appropriate given the current game context-without stating or implying the action.

Action Decision: Based on the scene understanding and inferred intent, choose the most reasonable next atomic action from the following valid action set:
NOOP, FIRE, UP, RIGHT, LEFT, DOWN, UPRIGHT, UPLEFT, DOWNRIGHT, DOWNLEFT, UPFIRE, RIGHTFIRE, LEFTFIRE, DOWNFIRE, UPRIGHTFIRE, UPLEFTFIRE, DOWNRIGHTFIRE, DOWNLEFTFIRE.

Instructions:
- Base your reasoning on multiple frames to capture temporal patterns and object interactions.
- Ensure your decision aligns with inferred goals and basic game logic.
- Wrap your reasoning in <think>...</think> tags.
- Wrap your final action decision in <answer>...</answer> tags, containing only the action name.
\end{lstlisting}
\end{center}

Problem prompt:

\begin{center}
\begin{lstlisting}[style=prompt]
You are given a sequence of consecutive video game frames (observations).

Your task is to analyze the visual scene and reason about the game's dynamics to infer the most appropriate next action.

- Provide your reasoning based Scene Analysis and Game Reasoning dimensions, inside `<think>...</think>` tags.
- Then, output the **final Action Decision** inside `<answer>...</answer>` tags.

Choose the action from the following valid set:
`NOOP`, `FIRE`, `UP`, `RIGHT`, `LEFT`, `DOWN`, `UPRIGHT`, `UPLEFT`, `DOWNRIGHT`, `DOWNLEFT`, `UPFIRE`, `RIGHTFIRE`, `LEFTFIRE`, `DOWNFIRE`, `UPRIGHTFIRE`, `UPLEFTFIRE`, `DOWNRIGHTFIRE`, `DOWNLEFTFIRE`.

\end{lstlisting}
\end{center}

Example answer:

\begin{center}
\begin{lstlisting}[style=prompt]
<think>Let me go with understanding the current state. The main game area spans from coordinates (0, 104) to (159, 171). I see
a player at (20, 137, 28, 174).
a ball at (27, 147, 31, 157).
a target at (133, 119, 135, 123).
a target at (129, 125, 131, 129).
a target at (125, 131, 127, 135).
a target at (133, 131, 135, 135).
a target at (121, 137, 123, 141).
a target at (129, 137, 131, 141).
a target at (125, 143, 127, 147).
a target at (133, 143, 135, 147).
a target at (129, 149, 131, 153).
a target at (133, 155, 135, 159).
a score at (38, 5, 125, 18).
a time at (29, 16, 61, 35).
Then, Let me consider what the optimal next action would be.
The game appears to be about targeting or hitting specific objects. The player is positioned on the left, with a collection of objects on the right side of the screen. The targets remain in the same position. There is a score and a timer. It seems the goal is to hit those targets within the given time to increase the score. The player is close to the baseline, and the relative positions of the player and the target have not changed. A change in the movement might not be optimal at this time.</think>
<answer>UPFIRE</answer>

\end{lstlisting}
\end{center}

\subsection{Hyperparemeters}

\begin{table}[H]
\centering
\small
\begin{tabular}{ll}
\toprule
\textbf{Hyperparameter} & \textbf{Value} \\
\midrule
Learning rate & $2\times 10^{-5}$ \\
Training epochs & 10 \\
Max sequence length & 4096 \\
Per-device batch size & 16 \\
Gradient accumulation steps & 4 \\
Precision & bf16 \\
Packing & Enabled \\
Gradient checkpointing & Enabled \\
Distributed training & ZeRO-3 (via Accelerate) \\
\bottomrule
\end{tabular}
\caption{Core hyperparameters used for supervised fine-tuning.}
\label{tab:sft_hparams}
\end{table}

\section{Meta-reasoning}

\subsection{Hyperparameters}
\begin{table}[h]
\centering
\small
\begin{tabular}{ll}
\toprule
\textbf{Hyperparameter} & \textbf{Value} \\
\midrule
Memory capacity & 20 \\
Max reasoning tokens & 1024 \\
Initial interval $k_{\text{init}}$ & 3 \\
Minimum interval $k_{\text{min}}$ & 3 \\
Maximum interval $k_{\text{max}}$ & 15 \\
Meta-reasoning model & gpt-4o-mini \\
\bottomrule
\end{tabular}
\caption{Key hyperparameters for the Meta-Reasoning Module.}
\label{tab:meta_hparams}
\end{table}

\subsection{Prompt}

\begin{center}
\begin{lstlisting}[style=prompt]
You are managing a memory system of rule (current: {current_count}/{max_capacity} rule).
Analyze the recent trajectory and current rules to decide memory operations.
Return meta reasoning process in <meta>reason about the memory operations you are going to make</meta>
followed by arbitraty number of operations using these tags:
<add> Some knowledge about the game mechanics, or good strategies summarised from the experience with reasoning </add> - to add a new rule (if space available)
<delete>rule_id</delete> - to remove rule by ID (0-based index)
<keep/> - to make no changes

rules should be the game mechanisms and/or strategies learned from the experience.
Avoid duplicates and contradictions.
To update a rule, use <delete>old_id</delete> then <add>new rule</add>.
If no changes needed, use <keep/>.
\end{lstlisting}
\end{center}

\section{Metacognive Test-time Reinforcement Learning}
\subsection{Hyperparameters}

We build our test-time reinforcement learning pipeline on top of the VERL framework \citep{sheng2024hybridflow}, leveraging its GRPO training pipeline while integrating our self-implemented TTRL reward mechanism.

\begin{table}[h]
\centering
\small
\begin{tabular}{ll}
\toprule
\textbf{Hyperparameter} & \textbf{Value} \\
\midrule
Reward manager &  TTRL \\
Algorithm & GRPO \\
LoRA rank & 64 \\
LoRA $\alpha$ & 32 \\
LoRA target modules & all linear layers (non-visual) \\
Max prompt length & 2048 \\
Max response length & 2048 \\
Train batch size & 8 \\
PPO mini-batch size & 8 \\
Micro-batch size / GPU & 4 \\
Learning rate & $3\times 10^{-6}$ \\
Rollout group size $n$ & 8 \\
Samples per prompt (reward) & 2 \\
Entropy coefficient & 0 \\
KL loss in actor & disabled \\
KL in reward & disabled \\
Total epochs & 5 \\
\bottomrule
\end{tabular}
\caption{Key hyperparameters used for test-time reinforcement learning with MCT-RL.}
\label{tab:ttrl_hparams}
\end{table}



\end{document}